\pdfoutput=1
\documentclass{article}
\usepackage[preprint]{corl_2026}
\usepackage[utf8]{inputenc}
\usepackage[T1]{fontenc}
\usepackage{url,booktabs,amsfonts,amsmath,nicefrac,microtype,xcolor,graphicx}
\usepackage{placeins}

\newcommand{\method}{\textsc{Encore}}

\usepackage[most]{tcolorbox}
\DeclareFontFamily{T1}{hanken}{}
\DeclareFontShape{T1}{hanken}{m}{n}{<-> ./fonts/hanken-r}{}
\DeclareFontShape{T1}{hanken}{b}{n}{<-> ./fonts/hanken-b}{}
\DeclareFontShape{T1}{hanken}{bx}{n}{<-> ssub * hanken/b/n}{}
\pdfmapline{+hanken-r < ./fonts/HankenGrotesk-Regular.ttf <T1-WGL4.enc}
\pdfmapline{+hanken-b < ./fonts/HankenGrotesk-Bold.ttf <T1-WGL4.enc}
\DisableLigatures[f]{encoding=T1,family=hanken}   
\newcommand{\hanken}{\fontfamily{hanken}\selectfont}
\definecolor{titlebg}{HTML}{F1F1FD}
\definecolor{titlefg}{HTML}{1B1E22}
\definecolor{brandnavy}{HTML}{090943}
\definecolor{brandblue}{HTML}{2E2AFF}
\usepackage{listings}
\tcbuselibrary{listings,breakable}
\definecolor{codebg}{HTML}{FAF9F6}
\definecolor{codeframe}{HTML}{D3D0C7}
\definecolor{codeink}{HTML}{1B1E22}
\definecolor{codekw}{HTML}{6367BD}
\definecolor{codestr}{HTML}{9A6A12}
\definecolor{codecm}{HTML}{6F757C}
\lstdefinestyle{encore}{language=Python,
  basicstyle=\ttfamily\scriptsize\color{codeink},   
  keywordstyle=\color{codekw}, stringstyle=\color{codestr}, commentstyle=\color{codecm},
  morecomment=[s]{"""}{"""},
  numbers=left, numberstyle=\ttfamily\tiny\color{codecm}, numbersep=6pt,
  columns=fullflexible, keepspaces=true, tabsize=4, showstringspaces=false, upquote=true,
  breaklines=true, postbreak=\mbox{\textcolor{codecm}{$\hookrightarrow$}\space}}
\newtcbinputlisting{\codefile}[2]{listing file={#2}, listing only,
  listing options={style=encore}, enhanced, breakable,
  colback=codebg, colframe=codeframe, boxrule=0.5pt, arc=5pt,
  left=18pt, right=6pt, top=4pt, bottom=4pt,
  colbacktitle=codebg, coltitle=codeink, titlerule=0.5pt, toptitle=3pt, bottomtitle=3pt,
  title={\hanken\small #1}}
\hypersetup{pdftitle={Encore: Few-Shot Agentic Discovery of Manipulation Strategies},
  pdfauthor={Yifan Kang, Zihan Wang, Zhiwen Fan, Bangya Liu}}

\begin{document}
\thispagestyle{empty}
\begin{tcolorbox}[enhanced, frame hidden, colback=titlebg, arc=10pt, before skip=0pt,
  left=0.5cm, right=0.5cm, top=0.5cm, bottom=0.5cm, grow to left by=1.5pt, grow to right by=1.5pt,
  overlay={\node[anchor=south east, xshift=-0.5cm, yshift=0.5cm] at (frame.south east)
    {\includegraphics[width=2.2cm]{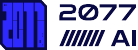}};}]
  \setlength{\parindent}{0pt}
  {\raggedright
    {\hanken\bfseries\fontsize{19}{25}\selectfont\color{brandnavy}
      Encore: Few-Shot Agentic Discovery\\of Manipulation Strategies\par}
    \vskip 0.25cm
    {\hanken\bfseries Yifan Kang$^{1,2}$,\enspace Zihan Wang$^{2}$,\enspace
      Zhiwen Fan$^{3}$,\enspace Bangya Liu$^{2,4,\dagger}$\par}
    \vskip 0.15cm
    \mbox{$^1$Massachusetts Institute of Technology,}\hskip 0.5em
    \mbox{$^2$2077AI Open Source Foundation,}\hskip 0.5em
    \mbox{$^3$Texas A\&M University,}\hskip 0.5em \mbox{$^4$University of Wisconsin--Madison}\par
    {\small $^\dagger$Corresponding author\par}}
  \vskip 0.5cm
  \color{titlefg}
Coding agents can now write, run, and debug programs with little human help.
Robot tasks, however, are usually specified by a sentence that leaves out how
to grasp, in what order to make contact, and what the result should look
like, and an agent given only the sentence must find these details by trial
and error. We introduce \method{}, which gives the agent a few
demonstrations as evidence to read rather than as training data. A
deterministic builder distills each demonstration into a pack of multi-view
keyframes, gripper events, frame strips, and the full trajectory. A coding
agent studies the pack, writes a policy program against a fixed perception
and action API, refines it iteratively over a few development rollouts, and freezes it before a sealed evaluation that never reveals the
success signal. On LIBERO-PRO, the agent's first program already succeeds in
half of the perturbed tasks with demonstrations and in one task without them, and the
frozen programs outperform the strongest prior agentic system run with the
same language model (96.3\% against 89.3\%). On RoboDojo tasks whose
instructions leave the goal unstated, no program succeeds without
demonstrations. \method{} also runs on a real bimanual robot, learning cube
handover and cup inversion from five demonstrations each.
  \vskip 0.5cm
  {\small\hypersetup{urlcolor=brandblue}\setlength{\parskip}{1pt}
    {\hanken\bfseries Code:} \href{https://github.com/YIFANK/encore}{\hanken github.com/YIFANK/encore}\par
    {\hanken\bfseries Project page:} \href{https://yifank.github.io/encore/}{\hanken yifank.github.io/encore}\par}
\end{tcolorbox}
\vskip 0.38cm

\begin{figure}[b]
  \centering
  \includegraphics[width=\textwidth]{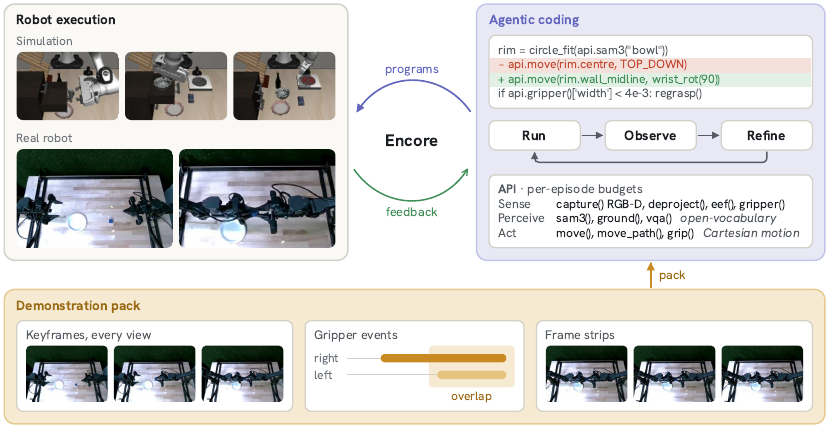}
  \caption{\method{}. A coding agent writes a policy program, runs it in
  simulation or on a bimanual robot, and refines it iteratively, with no
  success signal at run time. In the
  handover pack shown, the two grips overlap through the transfer, which
  the task sentence never mentions.}
  \label{fig:teaser}
\end{figure}

\section{Introduction}

Consider how an engineer would program the bimanual cube handover in
Figure~\ref{fig:teaser} from five demonstrations. They would notice that the
giving arm presents the cube from the side, that the receiving gripper rolls a
quarter turn so the finger planes cross, and that it closes before the giving
gripper opens. The task sentence mentions none of this, and each detail is
visible in only a few frames. The engineer would then write a program that
reproduces the sequence in a new layout, run it, and repair it. Learning a
policy from the same five trajectories remains
difficult~\citep{johns2021coarse,dreczkowski2025thousand}: they cover a tiny
part of the state space, and without synthetic
augmentation~\citep{mandlekar2023mimicgen,xue2025demogen} a policy trained on
them fails as soon as it leaves the demonstrated path.

The engineer's workflow of writing, running, and repairing a program is
exactly what coding agents now do well. Given tools, an execution
environment, and a budget of attempts, a frontier language model can write a
program, run it, and repair it from the result. Code-as-policies systems have already shown that robot
policies can be written as programs over perception and control
APIs~\citep{liang2023code,singh2023progprompt}, while more recent agents can
debug failed rollouts and accumulate reusable skills across
tasks~\citep{wang2023voyager,aspire2026,zhang2026rats}.

Coding agents suggest another use for demonstrations. Modern agents
interpret images and text and write analysis code to inspect unfamiliar
data, so we convert each teleoperated demonstration into a structured record
an agent can query: keyframes at the demonstrator's gripper actions, frame
strips around each contact event, and the complete underlying trajectory.
Such a record states both the intended outcome and the strategy that reached
it: where the demonstrator grasped, at what orientation, and along which
path.

Existing approaches do not provide demonstrations to coding agents in a form
they can use effectively. A teleoperated episode is a dense multimodal stream
containing hundreds of timesteps, several camera views, and many joint-state
channels. The moments that determine the outcome occupy a small fraction of the
frames.
Compressing the episode into a textual summary~\citep{wang2023demo2code} or a
compact token sequence~\citep{dipalo2024kat} requires deciding in advance
which details matter and may discard unexpected but essential cues, such as a
change in grasp orientation or a wobble before a regrasp. Placing the entire
episode in context creates the opposite problem: it overwhelms the agent with
incidental details. In our experiments, handing even the strongest prior
agentic pipeline~\citep{aspire2026} the same demonstrations as raw trajectory
files brought it no measurable gain.

We introduce \method{}, an agentic framework for few-shot robot manipulation.
A deterministic builder first transforms each demonstration into the
evidence described above. A fresh coding agent then explores this
evidence using analysis code that it writes itself, infers the task, and
implements a policy against a narrow perception and action API. The agent then refines the policy iteratively over a small number of development rollouts. To measure the demonstrations'
contribution, we compare three demonstrations ($K{=}3$) with none ($K{=}0$)
under the same model, API, development budget, and evaluation protocol. The design places the
boundary between agent and policy at development time: the language model
writes, tests, and repairs the program, and the deployed policy makes no
model calls, so it can be evaluated on thousands of sealed episodes with no
agent in the loop.

Our central claim is that demonstrations act as a search prior for agentic
policy construction: they expose manipulation strategies that the agent can
transfer to new layouts, compose under new instructions, and turn into
executable programs. On LIBERO-PRO~\citep{zhou2025liberopro}, the first
program written with demonstrations already succeeds during development in
about half of the perturbed cells, and the first program written without them
almost never does; given enough attempts, the agent without demonstrations
reaches a similar final rate (Figure~\ref{fig:deveff}). Where the agent's own
search fails, the demonstrations can supply the missing strategy, and in one
cell the demonstrated grasp decides every episode (Section~\ref{sec:task}). On
RoboDojo, whose sentences leave the goal or the contact strategy unstated, no
program succeeds without demonstrations. Our contributions are:
(i) a distiller that turns teleoperated demonstrations into evidence a coding
agent can query without being told what it means; (ii) a harness that
develops, verifies, and freezes policy programs with no access to the success
signal, whose programs succeed on 96.3\% of sealed LIBERO-PRO perturbation
episodes; and (iii) a controlled comparison on LIBERO-PRO, 22 RoboDojo tasks,
and a real bimanual robot that shows where demonstrations change what the
agent discovers.

\section{Related work}

\subsection{Code as policies and agentic robotics}

Language models can express robot policies as code over perception and
control primitives~\citep{liang2023code,singh2023progprompt}, and successors
structure that interface with spatial value maps~\citep{huang2023voxposer} or
relational keypoint constraints~\citep{huang2024rekep}. A second line of work
makes the model an agent: it executes its programs, reads the outcome, and
revises them, accumulating reusable skills across
tasks~\citep{aspire2026,fu2026capx,zhang2026rats,tsui2026faea}.
ASPIRE~\citep{aspire2026} is the strongest published system of this kind on
LIBERO-PRO~\citep{zhou2025liberopro} and our primary comparison. It iterates
from language alone, and its execution traces are annotated with
human-written diagnostic rules, including the gripper width that counts as a
grasp and the width that counts as closing on air. Our harness gives no such aids. The brief states only the task, traces are
passed on uninterpreted, and each agent builds its own diagnostics.

\subsection{Learning from few demonstrations}

Demonstration-conditioned code generation compresses trajectories into
prompts, by recursive summarization~\citep{wang2023demo2code} or keypoint
action tokens~\citep{dipalo2024kat}. The compression fixes in advance what
the model may see, and generation is single-pass. Few-shot imitation instead
transfers the demonstrated trajectories
directly~\citep{johns2021coarse,dipalo2024dinobot,vosylius2025instant}, and
demonstration-augmentation systems synthesize training data from a few
seeds~\citep{mandlekar2023mimicgen,garrett2024skillmimicgen,xue2025demogen};
in both cases the resulting artifact is a network whose behavior cannot be
read, audited, or edited. Unlike these methods, \method{} exposes distilled but uninterpreted
demonstration data to an iterative coding agent and evaluates the resulting
frozen program on held-out states.

\section{Method}
\label{sec:sys}

\paragraph{Problem statement.} A task is specified by one sentence of intent
and $K$ recorded human demonstrations on the same embodiment: in simulation
three episodes drawn from the benchmark's own demonstration files, on
hardware five that we teleoperate. The output is a \emph{policy program}: a
module that runs in a process holding no simulator and no robot handle, and
that acts on the world through a fixed API. Evaluation uses held-out initial states, and the success predicate is never exposed to the agent or to the program. The agent may run a bounded number of
development episodes on a disjoint band of states and has no access to asset files or object poses.

\method{} has three components (Figure~\ref{fig:teaser}): a
\emph{demonstration distiller} (\S\ref{sec:packs}), a
\emph{perception-and-action API} identical in simulation and on hardware
(\S\ref{sec:api}), and a \emph{development loop} (\S\ref{sec:loop}) that
ends with the program frozen for the sealed evaluation of
Section~\ref{sec:exp}. Tasks share no code.

\subsection{Demonstration distiller}
\label{sec:packs}

A demonstration is a stream $D=\{(I_t, s_t, a_t)\}_{t=1}^{T}$, where $T$ is
the number of timesteps and $I_t$ collects every camera view. The state
$s_t=(p_t, q_t, g_t)$ carries each arm's 6-DoF end-effector pose, joints,
and gripper width. A handful of the $T$ steps decide the task, so
distillation must select. Our distiller selects only on signals the
demonstrator produced and does not say what they mean: a builder that tags a
frame \emph{release} has already done part of the work we want the agent to
do. The pack of a demonstration is
\[
\mathcal{P}(D) \;=\; \big(\ell,\; K,\; E,\; \tau,\; A\big).
\]
$\ell$ is the intent sentence. $K=\{(t, I_t, s_t): t \in T_K\}$ is the
keyframe set: every camera view and the full state at each selected
instant. The indices $T_K$ are the demonstrator's own annotations, gripper
sign transitions $\{t: \operatorname{sign} g_t \neq \operatorname{sign}
g_{t-1}\}$ joined with the sharpest heading breaks of the smoothed
end-effector velocity. A multi-stage task therefore arrives pre-segmented,
with no video understanding in the pipeline. $E$ is the gripper-event
table: each open and close, with its time, arm, and width signal. The
labels describe the width signal only, and the agent decides which opening
is a \emph{release}. On hardware each event also carries a frame strip, the tiled frames spanning
roughly $1.5$\,s around it.
$\tau$ is the state path at stride five, units declared. $A$ is the
complete action stream. A whitelist validator removes all other fields, including simulator state,
object poses, and file paths, so the pack cannot lead the agent to benchmark
files.

\subsection{Perception-and-action API}
\label{sec:api}

The API draws the line between what we give the agent and what it must work
out. It supplies sensing and motion, the same in simulation and on hardware
(Table~\ref{tab:api}). Everything specific to the task, including which object
to take, how to grasp it, where it goes, and how to tell that it got there,
must come from the pack or from what the agent measures during development. Perception and
motion calls draw on a per-episode budget, so a program cannot buy success
with unlimited queries. On the robot the rig also serves SAM segmentation and
a guarded top-down pick and place; the handover program calls neither.

\begin{table}[t]
\centering
\caption{The API a policy program calls, the same in simulation and on
hardware.}
\label{tab:api}
\footnotesize
\begin{tabular}{lll}
\toprule
 & call & returns \\
\midrule
sense & \texttt{capture(cam)} & RGB-D frame with intrinsics and extrinsics \\
 & \texttt{deproject(u, v)} & 3-D point in the robot frame \\
 & \texttt{eef()}, \texttt{tool\_rotation()}, \texttt{gripper()} & pose, tool rotation, jaw width and effort \\
perceive & \texttt{sam3(q)}, \texttt{ground(q)}, \texttt{vqa(q)} & masks, boxes, or an answer for a text prompt \\
act & \texttt{move(xyz, R)}, \texttt{move\_path(pts, R)} & Cartesian motion; returns the position residual \\
 & \texttt{grip(w)} & sets the jaw width \\
\bottomrule
\end{tabular}
\end{table}

\textbf{Example.} The frozen drawer program of Section~\ref{sec:task}
(Appendix~\ref{app:programs}) turns a \texttt{capture} into a point cloud,
finds the cabinet face, and takes the lowest handle bar standing off it. It
tilts the open gripper $20^\circ$ below horizontal, the attitude it read from
the demonstrated pulls, seats the fingertips on top of the bar, and drags
with a \texttt{move} target far out and low, as the demonstrated actions do.
It then captures again and drags a second time if the drawer front has come
out less than 12\,cm.

\subsection{Development loop}
\label{sec:loop}

Each task gets one fresh agent, which explores the pack with analysis code
it writes itself, implements a program, and revises it over development
episodes on a disjoint band of initial states. Because benchmark success is unavailable at runtime, and evaluation
statically refuses any program that reads the episode termination flag,
programs implement their own grasp and placement checks using
proprioception and RGB-D. When development ends the agent selects one version and freezes it.

\section{Experiments}
\label{sec:exp}

\subsection{Setup}
\label{sec:setup}

\textbf{Benchmark.} LIBERO-PRO~\citep{zhou2025liberopro} perturbs
LIBERO~\citep{liu2023libero} along a layout axis (\textsc{Pos}), which
relocates objects and the fixtures they sit on, and an instruction axis
(\textsc{Task}), which re-authors the sentence and with it the rewarded end
state. Neither perturbed axis carries demonstrations of its own. We evaluate
all 30 tasks of the spatial, object, and goal suites on all three axes: 90
cells, 50 sealed episodes each. On the 30 unperturbed cells \method{} reaches
98.6\% (Table~\ref{tab:pertask}); Sections~\ref{sec:pos} and~\ref{sec:task}
cover the two perturbed axes. Every cell runs the loop of
Section~\ref{sec:loop} once, with a fresh opus-5 agent that shares no files
with any other cell. It develops on 15 initial states, and one frozen program
is evaluated once on 50 sealed states and scored by the benchmark's own
predicate (Appendix~\ref{app:protocol}).

\textbf{Baselines.} We compare against the results \citet{aspire2026} report
for the VLA baselines ($\pi_{0.5}$; OpenVLA and $\pi_0$ score zero and are
omitted), for their code-as-policies agent without a skill library
(CaP-Agent0), and for ASPIRE on opus-4.6. Because these come from an older
model, we also rerun ASPIRE's released code at a pinned commit on opus-5 over
the same 60 perturbation cells (Appendix~\ref{app:protocol}). The two
harnesses give their agents different tools (Appendix~\ref{app:contracts}).

\subsection{Transferring a demonstrated strategy to a new layout}
\label{sec:pos}

\begin{figure}[!h]
  \centering
  \includegraphics[width=0.88\textwidth]{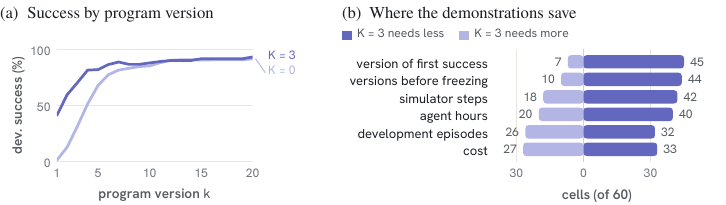}
  \caption{Development on LIBERO-PRO, \textsc{Pos} and \textsc{Task} cells
  pooled. (a) Mean development success of each cell's current program at
  version $k$ (60 cells; a frozen cell keeps its final version). (b) Paired per cell, how often the $K{=}3$ agent
  needs less or more of each resource. The demonstrations reduce the number of attempts, while agent time and cost
  stay similar (0.51 against 0.55 hours per cell), because the $K{=}3$ agent
  spends part of the saved iterations reading the pack. Same harness, budget, model, and sealed evaluation.}
  \label{fig:deveff}
\end{figure}

On the \textsc{Pos} axis the sentence is unchanged and the objects and their
fixtures have moved, so the question is whether the agent can carry a
demonstrated strategy to a layout it has not seen. It usually can, and from
its first program. The first program written with three demonstrations
already succeeds on some development state in 11 of 30 cells, and on average
on 29\% of its development episodes; the first program written without them
succeeds nowhere. The first success arrives at a median of version 2 against
4, earlier in 21 cells and later in 5 (Figure~\ref{fig:deveff}). The final rates are close, 1454 and 1437 of 1500 sealed episodes: with
15 development episodes the $K{=}0$ agent measures most of
what the pack states, including the grasp geometry and the goal site.

\subsection{Composing demonstrated skills under a new instruction}
\label{sec:task}

\begin{figure}[!h]
  \centering
  \includegraphics[width=\textwidth]{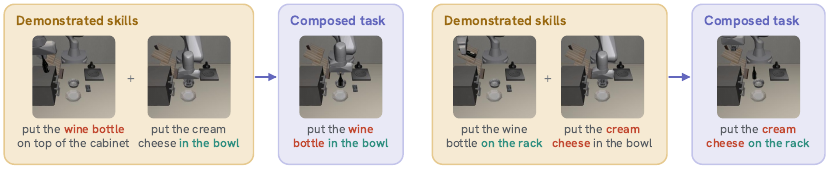}
  \caption{Recombination on the \textsc{Task} axis. Each cell gets two packs
  (left, centre) and a sentence asking for a third task (right) that neither
  demonstrates; red marks the object the sentence takes, teal the target,
  and which pack supplies which half flips between the two cells. Both cells
  score $50/50$ with the two packs.}
  \label{fig:crossskill}
\end{figure}

The \textsc{Task} axis rewrites the sentence and the rewarded end state and
keeps the scene. The rewrites are recombinations: each cell receives two
packs, neither of which demonstrates its goal, and the agent must decide from
the sentence which part of each pack to reuse (Figure~\ref{fig:crossskill}).
Here the demonstrations help more. The first program already succeeds in 18
of 30 cells, on average on 53\% of its development episodes, against one cell
and 2\% without demonstrations, and the median first success arrives at
version 1 against 5. The frozen programs differ as well: paired per cell,
$K{=}3$ wins 10, loses 2, and ties 18, and succeeds on 1435 of 1500 sealed
episodes against 1409.

\textbf{The drawer cell.} On
\texttt{open\_middle\_drawer}-\textsc{Task} the sentence asks for the
bottom drawer, while the demonstrations open the middle one. The cell scores
$50/50$ with demonstrations and $0/50$ without, in both acquisitions. The
$K{=}0$ agent could open the other two drawers, but in 33 program versions
no grasp it tried engaged the bottom handle: tilted pinches, a hook from
below the bar, and pressing its closed jaws on top of the bar and dragging
outward, which did not move the drawer. It froze a program that opens the
middle and top drawers and concluded that the bottom handle cannot be
engaged through this API. The demonstrations of the middle drawer show the
missing piece: throughout the pull the demonstrator's actions drive
downward as well as outward, with the wrist tilted 15 to $36^\circ$ below
horizontal. The $K{=}3$ program seats its open fingertips on the bar and
reproduces that drag, and the bottom drawer opens
(Appendices~\ref{app:cases} and~\ref{app:programs}). A failed search during
development does not show that a task is infeasible, and a demonstration can
supply the strategy the search missed.

\textbf{The plate cell.} The largest loss runs the other way. On
\texttt{put\_bowl\_top\_cabinet}-\textsc{Task}, which asks for the plate on
top of the cabinet, the $K{=}3$ agent never held the plate long enough to
carry it, recorded that no grasp its jaws could keep would reach the
cabinet, and froze after 11 versions at $0/50$. The $K{=}0$ agent kept searching for 23 versions, found a grasp that lifts the
plate along an arc, and scored $50/50$. The demonstrations shorten the
search; they do not guarantee that it ends in the right place.

\begin{figure}[t]
  \centering
  \includegraphics[width=0.88\textwidth]{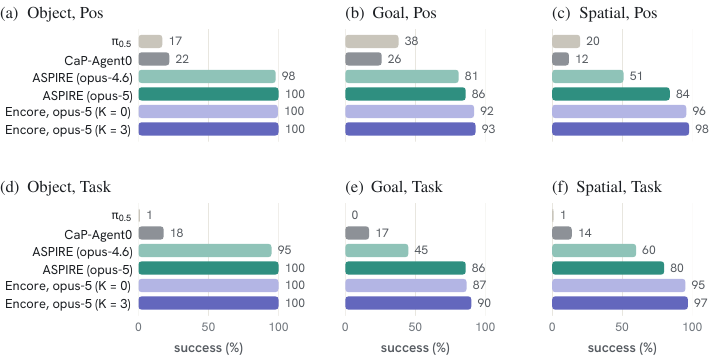}
  \caption{LIBERO-PRO~\citep{zhou2025liberopro}, sealed success per suite
  and axis (10 tasks $\times$ 50 episodes). $\pi_{0.5}$, CaP-Agent0, and
  ASPIRE on opus-4.6 as reported by \citet{aspire2026}; ASPIRE on opus-5 is
  our rerun of its released code. $K{=}1$ is in Appendix~\ref{app:tables}.}
  \label{fig:bars}
\end{figure}

\subsection{Comparison with prior systems}
\label{sec:prior}

Over the 60 perturbation cells the $K{=}3$ programs succeed on 2889 of 3000
sealed episodes (96.3\%) and the $K{=}0$ programs on 2846 (94.9\%), against
89.3\% for ASPIRE rerun on the same model and 71.7\% as published; most of
the margin lies in the spatial suite (Figure~\ref{fig:bars}; per task in
Appendix~\ref{app:tables}). With one demonstration the programs succeed on
2791 (93.0\%). An earlier acquisition of the same cells, run before we
removed a shared note file from the brief, showed a $K{=}3$ advantage of 252
episodes. The clean protocol reproduces the $K{=}3$ number but not the gap,
so we do not report the gap as an effect. Removing the verification loop
while keeping the demonstrations drops the goal suite from 440 to 117 of 450
(Appendix~\ref{app:abl}).

\subsection{Underspecified tasks: RoboDojo}
\label{sec:rd}

\begin{figure}[t]
  \centering
  \includegraphics[width=\textwidth]{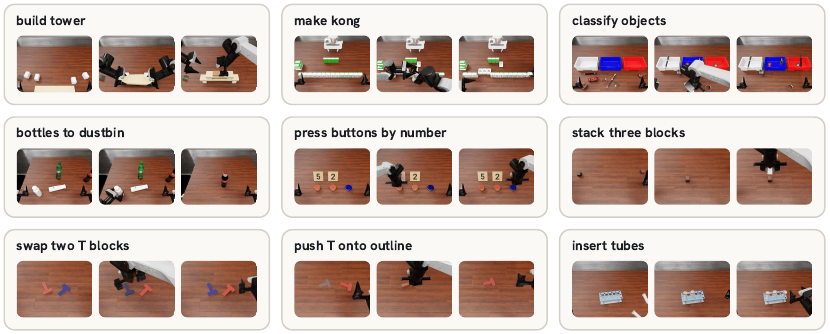}
  \caption{RoboDojo tasks, one sealed $K{=}3$ success each (start, middle,
  end). The first row states its goal only through the demonstrations; the
  others need a delicate contact or a visual cue.}
  \label{fig:rdtasks}
\end{figure}

RoboDojo~\citep{robodojo2026} is a bimanual Isaac Sim benchmark. Its
instructions often leave the goal or the contact strategy unstated (``Build
a tower using the wooden blocks and wooden boards'', ``put the bottles into
the dustbin, using handover when needed''), while its judge scores a specific
structure or sequence. We ran the unchanged loop on 22 of its tasks
with three teleoperated demonstrations per task, a development band of
15 layouts, and a sealed band drawn from the official evaluation
layouts and scored by the benchmark's own judge (Figure~\ref{fig:rdtasks}, Table~\ref{tab:rd}). On 12 of
them, chosen to separate goal-underspecified from contact-sensitive tasks, we
add an arm that receives the same three demonstrations reduced to their keyframe
images, with every pose, gripper reading, and action removed.

\textbf{Without demonstrations, no program succeeds.} On the first 10 tasks
$K{=}3$ succeeds on 95 of 450 sealed episodes and $K{=}0$ on none. The 95
come from four tasks, and on each of them the $K{=}0$ agent solved the
manipulation but could not tell what to build. One wrote: ``the manipulation
is solved; the target structure is not known.'' The $K{=}3$ agents read the
structure from the keyframe images. With one demonstration the 10 tasks
reach 40 of 450.

\textbf{Images alone are not enough when the contact is delicate.} On
stacking, swapping, inserting tubes, and pushing the T, the images-only arm
falls well below the full pack, and on stacking and swapping it scores below
no pack at all: the agent saw a picture of the demonstrated grasp without its
measurements and spent its budget reconstructing them. Only on pressing
buttons by number, where the goal is visual, do the images carry the whole
effect; on tic-tac-toe they do best, because the full-pack agent kept the
demonstrations' pauses between turns and ran out of steps. Six
contact-precise tasks (plugging a charger, inserting a key, fastening screws,
making toast, storing a laptop, hanging mugs) score zero in every arm, which
marks the current limit of these programs.

\begin{table}[!t]
\centering
\caption{RoboDojo, successes in 50 sealed episodes per task, scored by the
benchmark's judge. Left: the tasks not shown (organize table, pack objects,
imitate sorting, fold clothes) score 0 in every arm, and one task has no
demonstration data. Right: tasks with an images-only arm; the six
contact-precise tasks score 0 in every arm ($K{=}3$, images and $K{=}0$ on 20
episodes). Bold: best arm.}
\label{tab:rd}
\scriptsize\setlength{\tabcolsep}{3.5pt}
\begin{tabular}{lccc}
\toprule
task & $K{=}3$ & $K{=}1$ & $K{=}0$ \\
\midrule
build tower & $\mathbf{36}$ & $0$ & $0$ \\
make kong & $\mathbf{36}$ & $0$ & $0$ \\
bottles to dustbin & $\mathbf{15}$ & $2$ & $0$ \\
classify objects & $\mathbf{8}$ & $7$ & $0$ \\
arrange by number & $0$ & $\mathbf{31}$ & $0$ \\
\midrule
10 tasks & $\mathbf{95}$ & $40$ & $0$ \\
\bottomrule
\end{tabular}\hspace{0.6em}
\begin{tabular}{lcccc}
\toprule
task & $K{=}3$ & $K{=}1$ & images & $K{=}0$ \\
\midrule
press buttons & $\mathbf{50}$ & $\mathbf{50}$ & $\mathbf{50}$ & $0$ \\
stack three blocks & $\mathbf{43}$ & $22$ & $0$ & $10$ \\
swap two T blocks & $\mathbf{47}$ & $44$ & $15$ & $38$ \\
push T onto outline & $\mathbf{18}$ & $0$ & $8$ & $0$ \\
insert tubes & $14$ & $\mathbf{34}$ & $0$ & $2$ \\
tic-tac-toe & $5$ & $0$ & $\mathbf{21}$ & $4$ \\
six contact-precise & $0$ & $0$ & $0$ & $0$ \\
\bottomrule
\end{tabular}
\end{table}

\begin{figure}[!t]
  \centering
  \includegraphics[width=\textwidth]{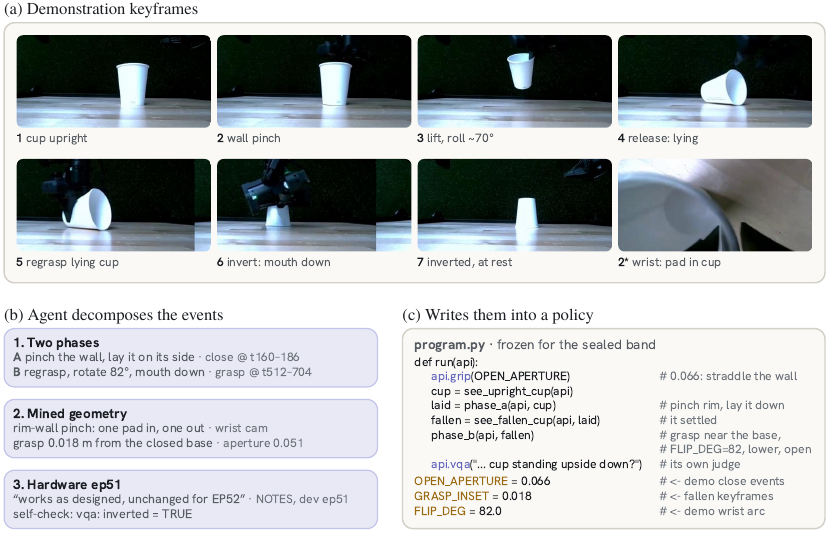}
  \caption{From five demonstrations to a frozen policy on the real robot:
  (a) a cup-inversion demonstration as the pack presents it; (b) the
  agent's reading, quoted from its notes; (c) the frozen program, which
  verifies its own outcome by querying a VLM about the scene.}
  \label{fig:flip}
  \vspace{8pt}
  \includegraphics[width=\textwidth]{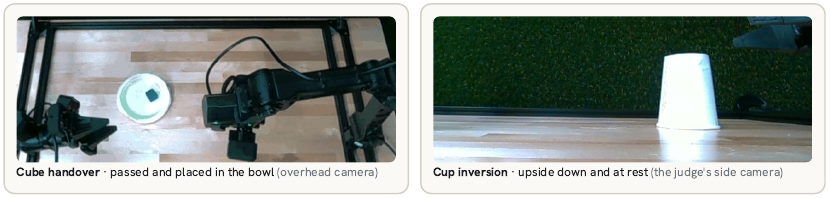}
  \caption{Goal states: the cube passed and placed in the bowl
  (overhead); the cup upside down and at rest (the judge's side camera).}
  \label{fig:rigtasks}
\end{figure}

\subsection{Real-robot experiments}
\label{sec:rig}

\paragraph{Setup.} We run the loop of Section~\ref{sec:loop} and the API of
Section~\ref{sec:api} unchanged on a
bimanual Trossen WidowX~AI: two 6-DoF arms with parallel jaws, observed by
four RealSense D405 cameras, the overhead one calibrated to the table
plane to $1.1$\,mm. We evaluate a two-arm cube handover and cup inversion
(Figure~\ref{fig:rigtasks}), five teleoperated demonstrations each,
distilled by the same builder as in simulation. A fresh agent develops from the pack and freezes one program; evaluation
initial states are operator-varied and recorded by the harness, and a
human who did not develop the program judges each trial.

\paragraph{Tasks.} The handover demonstrations
present the cube from the side, roll the receiving gripper a quarter turn
so the finger planes cross, and close the receiving grip before the giving
one opens. The cup demonstrations pinch the standing cup's wall, lay it on
its side, regrasp it near the base, and stand the wrist up so the opening
lands on the table, four gripper events rather than two (Figure~\ref{fig:flip}; Appendix~\ref{app:cases}).
On sealed 10-episode human-judged bands, both tasks score $9/10$.

\FloatBarrier
\bibliography{refs}

\appendix

\section{Main result tables}
\label{app:tables}

Table~\ref{tab:macro} reports the exact values visualized in
Figure~\ref{fig:bars}, and Table~\ref{tab:pertask} the per-task breakdown behind
them. All
\method{} numbers are 50-seed sealed evaluations (seeds 1--50) of one
frozen program per cell; baselines are as published
by~\citet{aspire2026}, except the opus-5 rerun of ASPIRE, which we ran at
a pinned commit (Section~\ref{sec:exp}). The \method{} $K{=}3$ and $K{=}0$ columns
are the clean acquisition of 2026-09-13 and the $K{=}1$ column its
counterpart of 2026-09-25; the earlier acquisition that included a shared
note file is not reported.

\begin{table}[h]
\centering
\caption{LIBERO-PRO success under position (\textsc{Pos}) and instruction
(\textsc{Task}) perturbation, macro-averaged over 10 tasks per suite.}
\label{tab:macro}
\small
\begin{tabular}{lcccccccc}
\toprule
 & \multicolumn{2}{c}{libero-object} & \multicolumn{2}{c}{libero-goal} & \multicolumn{2}{c}{libero-spatial} & \multicolumn{2}{c}{Overall} \\
\cmidrule(lr){2-3}\cmidrule(lr){4-5}\cmidrule(lr){6-7}\cmidrule(lr){8-9}
Method & Pos & Task & Pos & Task & Pos & Task & Pos & Task \\
\midrule
OpenVLA & 0.00 & 0.00 & 0.00 & 0.00 & 0.00 & 0.00 & 0.00 & 0.00 \\
$\pi_0$ & 0.00 & 0.00 & 0.00 & 0.00 & 0.00 & 0.00 & 0.00 & 0.00 \\
$\pi_{0.5}$ & 0.17 & 0.01 & 0.38 & 0.00 & 0.20 & 0.01 & 0.25 & 0.01 \\
CaP-Agent0 & 0.22 & 0.18 & 0.26 & 0.17 & 0.12 & 0.14 & 0.20 & 0.16 \\
ASPIRE (opus-4.6) & 0.98 & 0.95 & 0.81 & 0.45 & 0.51 & 0.60 & 0.77 & 0.67 \\
ASPIRE (opus-5, our rerun) & \textbf{1.00} & \textbf{1.00} & 0.86 & 0.86 & 0.84 & 0.80 & 0.90 & 0.89 \\
\method{} ($K{=}0$) & \textbf{1.00} & \textbf{1.00} & 0.92 & 0.87 & 0.96 & 0.95 & 0.96 & 0.94 \\
\method{} ($K{=}1$) & 0.99 & \textbf{1.00} & 0.89 & 0.85 & 0.90 & 0.94 & 0.93 & 0.93 \\
\method{} ($K{=}3$) & \textbf{1.00} & \textbf{1.00} & \textbf{0.93} & \textbf{0.90} & \textbf{0.98} & \textbf{0.97} & \textbf{0.97} & \textbf{0.96} \\
\bottomrule
\end{tabular}

\end{table}

\begin{table}[h]
\centering
\caption{Per-task success on LIBERO-PRO, 50 sealed episodes per cell.
\emph{Stock} is the unperturbed task. $K{=}3$ receives three demonstrations, $K{=}1$ one, $K{=}0$ none, with the framework otherwise identical.}
\label{tab:pertask}
\small
\begin{tabular}{lccccccc}
\toprule
 & Stock & \multicolumn{3}{c}{Pos} & \multicolumn{3}{c}{Task} \\
\cmidrule(lr){3-5}\cmidrule(lr){6-8}
Task & $K{=}3$ & $K{=}3$ & $K{=}1$ & $K{=}0$ & $K{=}3$ & $K{=}1$ & $K{=}0$ \\
\midrule
\multicolumn{8}{l}{\emph{libero-object}} \\
alphabet soup & 1.00 & 1.00 & 1.00 & 1.00 & 1.00 & 1.00 & 1.00 \\
bbq sauce & 1.00 & 1.00 & 1.00 & 1.00 & 1.00 & 1.00 & 1.00 \\
butter & 1.00 & 1.00 & 0.98 & 1.00 & 1.00 & 1.00 & 1.00 \\
chocolate pudding & 1.00 & 1.00 & 1.00 & 1.00 & 1.00 & 1.00 & 1.00 \\
cream cheese & 1.00 & 1.00 & 1.00 & 1.00 & 1.00 & 1.00 & 1.00 \\
ketchup & 1.00 & 1.00 & 1.00 & 1.00 & 1.00 & 1.00 & 1.00 \\
milk & 1.00 & 1.00 & 0.96 & 0.96 & 1.00 & 1.00 & 1.00 \\
orange juice & 1.00 & 1.00 & 1.00 & 1.00 & 1.00 & 1.00 & 1.00 \\
salad dressing & 1.00 & 1.00 & 1.00 & 1.00 & 1.00 & 1.00 & 1.00 \\
tomato sauce & 1.00 & 1.00 & 1.00 & 1.00 & 1.00 & 0.98 & 1.00 \\
\midrule
\multicolumn{8}{l}{\emph{libero-goal}} \\
open middle drawer & 0.96 & 0.70 & 0.06 & 0.44 & 1.00 & 0.00 & 0.00 \\
open top drawer put bowl & 0.98 & 1.00 & 0.98 & 1.00 & 1.00 & 1.00 & 0.98 \\
push plate front stove & 0.92 & 1.00 & 1.00 & 1.00 & 1.00 & 1.00 & 0.98 \\
put bowl on plate & 1.00 & 0.94 & 0.92 & 0.96 & 1.00 & 1.00 & 1.00 \\
put bowl on stove & 1.00 & 0.98 & 1.00 & 1.00 & 1.00 & 0.86 & 0.84 \\
put bowl top cabinet & 1.00 & 1.00 & 1.00 & 1.00 & 0.00 & 0.82 & 1.00 \\
put cream cheese in bowl & 1.00 & 0.92 & 0.98 & 1.00 & 1.00 & 0.94 & 0.88 \\
put wine on rack & 0.94 & 0.76 & 1.00 & 0.80 & 1.00 & 1.00 & 0.98 \\
put wine top cabinet & 0.98 & 1.00 & 1.00 & 1.00 & 1.00 & 1.00 & 1.00 \\
turn on stove & 1.00 & 1.00 & 1.00 & 1.00 & 1.00 & 0.88 & 1.00 \\
\midrule
\multicolumn{8}{l}{\emph{libero-spatial}} \\
bowl between & 0.96 & 0.98 & 1.00 & 1.00 & 0.92 & 0.80 & 0.92 \\
bowl cookie box & 1.00 & 0.92 & 0.38 & 1.00 & 1.00 & 1.00 & 1.00 \\
bowl next to plate & 0.96 & 1.00 & 1.00 & 1.00 & 1.00 & 1.00 & 1.00 \\
bowl next to ramekin & 1.00 & 0.98 & 0.98 & 1.00 & 0.98 & 0.96 & 0.92 \\
bowl on cookie box & 1.00 & 1.00 & 0.90 & 0.94 & 1.00 & 0.78 & 0.88 \\
bowl on ramekin & 1.00 & 1.00 & 1.00 & 0.90 & 1.00 & 1.00 & 1.00 \\
bowl on stove & 1.00 & 1.00 & 1.00 & 1.00 & 1.00 & 0.96 & 0.98 \\
bowl on wooden cabinet & 1.00 & 0.90 & 0.78 & 0.78 & 1.00 & 1.00 & 1.00 \\
bowl table center & 1.00 & 1.00 & 1.00 & 1.00 & 0.84 & 0.94 & 0.94 \\
bowl top drawer cabinet & 0.88 & 1.00 & 0.98 & 0.96 & 0.96 & 1.00 & 0.88 \\
\bottomrule
\end{tabular}

\end{table}
\FloatBarrier

\section{Development cost}
\label{app:cost}

Per cell, median over the sixty perturbation cells, clean acquisition: the
$K{=}3$ agent writes 4 program versions, runs 46.5 development episodes, and
spends 0.51 hours and 66k output tokens before freezing; the $K{=}0$ agent 6
versions, 53 episodes, 0.55 hours, and 69k tokens. Totals over the sixty
cells are 37 against 44 agent hours. The first version already succeeds on
some development state in 29 cells with demonstrations and in 1 without,
and the version of first success has median 2 against 4
(Figure~\ref{fig:deveff}). By comparison, ASPIRE's released loop on the same
model spends a median 0.99 hours and 91k output tokens per cell over the
sixty cells. A frozen program then costs nothing at evaluation time beyond
the simulator: a sealed LIBERO episode takes eleven seconds and no model call.

\FloatBarrier

\section{Additional ablations}
\label{app:abl}

\textbf{Verification loop and raw demonstrations.} Removing the
verification loop while keeping the demonstrations drops the goal suite from
440 to 117 of 450 (Table~\ref{tab:ablations}; both arms under the earlier
brief with the shared note file). Without the loop, the agent freezes programs it has not seen fail. Giving ASPIRE's loop the same three
demonstrations as raw trajectory files gains 6 episodes in 500,
within single-run variance. The demonstrations pay off when they are distilled
into a pack and the agent can test what it reads from them.

\textbf{When is the action channel necessary?} On RoboDojo the images-only
arm loses to the full pack where contact is delicate. To test whether this
holds for a class of fixtures and not only for particular cells, we ran the
three arms, under the earlier brief, on eight LIBERO-90 tasks with articulated fixtures (microwave,
three drawers, stove, opening and closing each). The other seven score
$50/50$ in every arm but one, which scores $42/50$. On \texttt{open\_bottom\_drawer} the full pack scores $50/50$ and
both no-action arms $0/50$. Two measurable conditions separate this fixture
from the seven: the bar protrudes less than the jaw's open half-width, so the
default pinch is geometrically impossible, and the pinch fails silently,
closing on air without a contact signature. On the microwave the pinch is
also impossible but fails loudly, the door prises the jaws open, and the
$K{=}0$ agent diagnosed its way to the hook. The demonstration was necessary only where the failure left no signal the agent could observe.

\begin{table}[h]
\centering
\caption{Ablations. Top: verification loop excised with demonstrations kept
(nine goal-suite tasks, 50 sealed episodes each; earlier brief with a note
file shared across cells). Bottom: ASPIRE's released
loop on opus-5 on the goal-swap axis (10 tasks $\times$ 50 seeds), with and
without the three demonstrations as raw trajectory files.}
\label{tab:ablations}
\footnotesize
\begin{tabular}{lc}
\toprule
Arm & Success \\
\midrule
\multicolumn{2}{l}{\emph{verification ablation, \method{}}} \\
\quad full system ($K{=}3$) & $440/450$ \\
\quad verification loop removed & $117/450$ \\
\midrule
\multicolumn{2}{l}{\emph{ASPIRE's loop}} \\
\quad as released & $431/500$ \\
\quad plus three demonstrations as raw trajectories & $437/500$ \\
\bottomrule
\end{tabular}
\end{table}

\section{Evaluation protocol}
\label{app:protocol}

Every cell runs the development loop once, headless, with a fresh agent that
has not seen any other task; the inner model is opus-5 for our arms and for
the rerun baseline. All agents receive the same brief. In the LIBERO-PRO and
RoboDojo experiments each cell is developed in isolation: agents share no
files with one another, and each agent derives the constants in its program
from its own pack and development runs. Development touches a disjoint band
of 15 initial states (seeds 51--65 on LIBERO-PRO). The selected program is
frozen, its hash recorded, and evaluated once by a coordinator on 50 sealed
states (seeds 1--50), with no per-episode result reaching any agent before the
band completes; the benchmark's own success predicate is the only judge. A
static check refuses programs that branch on the episode termination flag, so
each program must carry its own verification. Comparisons between arms are
paired per cell over the 60 matched perturbation cells. The ASPIRE rerun uses a
wrapper that restarts a session if it delivers a program before its
development sweep has finished; it reaches 89.3\%, against the 71.7\%
reported in the original paper. The two ablations of Appendix~\ref{app:abl}
were acquired earlier, under a brief that shared a note file across cells.

\section{Inputs available to each system}
\label{app:contracts}

The two contracts differ, and the difference
favors the baseline. ASPIRE's agents call a grasp planner and a
top-down grasp selector, fit oriented bounding boxes with a library routine,
read a coordinator-curated library of reusable \emph{code} at the start of
every session, and are handed diagnostic thresholds in the brief, down to the
gripper width that counts as a grasp and the width that counts as closing on
air. \method{}'s agents receive the demonstrations and none of these tools or hints,
and every threshold in a \method{} program was set by the agent that wrote it.

\section{Case studies}

\label{app:cases}

\paragraph{A demonstration is an existence proof (\textsc{Task}).}
(Section~\ref{sec:task}.) On \texttt{open\_middle\_drawer}-\textsc{Task} the
$K{=}0$ agent resolved the re-authored target correctly: it opened the middle
and top drawers on its development seeds, saw the benchmark reject both, and
turned to the bottom bar. Its notes record why every grasp failed. A side
pinch needs a horizontal wrist, and the horizontal hand extends 96\,mm below
the end effector while the bar stands 48\,mm above the table; rolling the
wrist reduces the overhang to 41\,mm, still too much. A hook must sit in the
slot behind the bar, which is 11\,mm deep, and the closed finger pair is
15\,mm thick. Pressing the closed jaws on top of the bar and dragging it
10\,cm outward left the drawer face where it was. Across 20 probe versions
and four wrist families, the deepest low pose it reached was a bar thickness
short of a grip. It wrote this down as a falsifiable claim that the handle
cannot be engaged, froze a program that opens the two drawers it can reach,
and scored $0/50$.

The $K{=}3$ agent hit the same wall with pinches and then read the pack: the
demonstrated pull keeps the wrist 15 to $36^\circ$ below horizontal, and its
actions stay saturated outward and downward (a commanded height change near
the maximum throughout the pull while the measured height does not change).
Seating the open fingertips on top of the bar and commanding a target far
out and low reproduces that pull; on its development seeds it dragged the
bottom drawer 19\,cm. The $K{=}0$ agent turned its failed attempts into a
claim about the robot, and the demonstrations show that the claim is false.

\FloatBarrier
\section{Frozen program listings}
\label{app:programs}

The two frozen programs of Section~\ref{sec:task}'s drawer cell, exactly as
evaluated (md5 \texttt{011b5af6} and \texttt{c2fcb7de} in the reproduction
bundle), less the \texttt{PROVENANCE} dictionary that our audit reads and the
policy never does. First the
$K{=}0$ program ($0/50$), then the $K{=}3$ one
($50/50$).

\subsection{Without demonstrations}
\codefile{\textbf{program\_k0.py} $\cdot$ frozen, $K{=}0$, $0/50$ sealed}{listings/program_k0.py}

\subsection{With demonstrations}
\codefile{\textbf{program\_k3.py} $\cdot$ frozen, $K{=}3$, $50/50$ sealed}{listings/program_k3.py}

\end{document}